\documentclass[]{template}

\usepackage[utf8]{inputenc}             
\usepackage[T1]{fontenc}                
\usepackage{url}                        
\usepackage{booktabs}                   
\usepackage{multirow}
\usepackage{colortbl}
\usepackage{multicol}
\usepackage{amsfonts}                   
\usepackage{nicefrac}                   
\usepackage{microtype}                  
\usepackage[dvipsnames]{xcolor}         

\usepackage{latexsym}

\usepackage{graphicx}
\usepackage{float}
\usepackage{subcaption}
\usepackage{wrapfig}
\usepackage{lipsum}
\usepackage{caption}

\usepackage{bm}

\usepackage{tabularx} 
\usepackage{ragged2e} 
\newcolumntype{L}{>{\RaggedRight\hangafter=1\hangindent=0em}X}

\usepackage{enumitem}

\usepackage{amsmath}
\usepackage{amssymb}
\usepackage{mathtools}
\usepackage{amsthm}

\setboolean{logo}{true}    

\usepackage[linesnumbered,ruled,vlined]{algorithm2e}

\hypersetup{
    colorlinks=true,
    linkcolor=red,
    citecolor=Cerulean,
    filecolor=magenta,      
    urlcolor=magenta,
}

\usepackage[capitalize,noabbrev]{cleveref}
\crefname{section}{§}{§§}
\Crefname{section}{§}{§§}

\usepackage{calligra}
\DeclareMathAlphabet{\mathcalligra}{T1}{calligra}{m}{n}

\usepackage{pifont}

\theoremstyle{plain}

\theoremstyle{definition}

\theoremstyle{remark}

\renewcommand{\paragraph}[1]{\vspace{1mm}\noindent\textbf{#1}}

\DeclareCaptionLabelFormat{cont}{#1~#2\alph{ContinuedFloat}}
\usepackage[most]{tcolorbox}
\tcbset{
  promptbox/.style={
    top=10pt,
    colback=lightgray!20,
    colframe=Black,
    colbacktitle=NavyBlue,
    enhanced,
    center,
    attach boxed title to top center={yshift=-0.1in,xshift=0.0in},
    boxed title style={boxrule=0pt,colframe=white,},
  }
}
\newtcolorbox{promptbox}[2][]{promptbox, title=#2,#1}
\tcbset{
  takeawaybox/.style={
    top=10pt,
    colback=lightgray!20,
    colframe=Black,
    colbacktitle=BurntOrange,
    enhanced,
    center,
    attach boxed title to top center={yshift=-0.1in,xshift=0.0in},
    boxed title style={boxrule=0pt,colframe=white,},
  }
}
\newtcolorbox{takeawaybox}[2][]{takeawaybox, title=#2,#1}
\tcbset{
  observationbox/.style={
    top=10pt,
    colback=lightgray!20,
    colframe=Black,
    colbacktitle=YellowGreen,
    enhanced,
    center,
    attach boxed title to top center={yshift=-0.1in,xshift=0.0in},
    boxed title style={boxrule=0pt,colframe=white,},
  }
}
\newtcolorbox{observationbox}[2][]{observationbox, title=#2,#1}

\usepackage{xspace}

\newcommand\blfootnote[1]{%
  \begingroup
  \renewcommand\thefootnote{}\footnote{#1}%
  \addtocounter{footnote}{-1}%
  \endgroup
}

\newcommand{\method}{CTRL-S\xspace}
\newcommand{\mytoprule}{
    \toprule
    \noalign{\vspace{-0.2mm}}
}
\newcommand{\mymidrule}{
    \noalign{\vspace{-0.8mm}}
    \midrule
    \noalign{\vspace{-1mm}}
}
\newcommand{\mybottomrule}{
    \noalign{\vspace{-0.6mm}}
    \bottomrule
}
\definecolor{oursgray}{gray}{0.95}
\definecolor{oursgreen}{rgb}{0.90, 0.96, 0.90}
\definecolor{oursblue}{rgb}{0.918, 0.941, 0.988}
\definecolor{improvementgreen}{rgb}{0.0, 0.4, 0.15}
\newcommand{\impv}[1]{\textcolor{improvementgreen}{\textsl{#1}}}

\usepackage{twemojis}
\usepackage{fontawesome}
\newcommand{\github}{\raisebox{-1.5pt}{\includegraphics[height=1.05em]{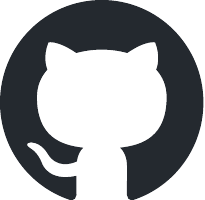}}\xspace}
\newcommand{\huggingface}{\raisebox{-1.5pt}{\includegraphics[height=1.05em]{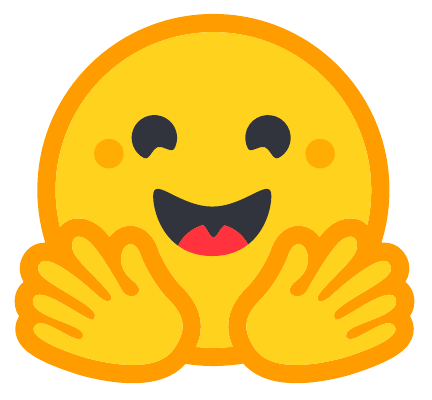}}\xspace}

\usepackage{CJK}

\title{Reliable Reasoning in SVG-LLMs via Multi-Task Multi-Reward Reinforcement Learning}

\author[1,2$*$]{Haomin Wang}
\author[2,3$*$]{Qi Wei}
\author[1,2]{Qianli Ma}
\author[2,4]{Shengyuan Ding}
\author[2,3]{Jinhui Yin}
\author[2]{Kai Chen}
\author[2$\dagger$]{Hongjie Zhang}

\affil[1]{Shanghai Jiao Tong University}
\affil[2]{Shanghai AI Laboratory}
\affil[3]{Nanjing University}
\affil[4]{Fudan University}

\begin{abstract}
With the rapid advancement of vision–language models, an increasing number of studies have explored their potential for SVG generation tasks. Although existing approaches improve performance by constructing large-scale SVG datasets and introducing SVG-specific tokens, they still suffer from limited generalization, redundant paths in code outputs, and a lack of explicit reasoning. In this work, we present \textbf{\method} (\textbf{C}hain-of-\textbf{T}hought \textbf{R}einforcement \textbf{L}earning for \textbf{S}VG), a unified framework that introduces a chain-of-thought mechanism to explicitly expose the model’s reasoning process during SVG generation. To support this structured reasoning, we construct \textbf{SVG-Sophia}, a high-quality dataset containing 145K samples across SVG code refinement, Text-to-SVG, and Image-to-SVG tasks. By training the model to generate group-level structured SVG code, \method significantly improves structural coherence and visual fidelity. Furthermore, we adopt the GRPO algorithm and design a multi-reward optimization framework, incorporating DINO, image–text similarity, format, and code efficiency rewards. Through joint multi-reward optimization and multi-task training, our approach systematically enhances overall generation capabilities. Extensive experiments show that \method outperforms existing methods, achieving higher task success rates, superior SVG code quality, and exceptional visual fidelity.
\end{abstract}

\begin{document}

\blfootnote{$*$ Equal Contribution: kiyotakawang@sjtu.edu.cn, qiwei@smail.nju.edu.cn}
\blfootnote{$\dagger$ Corresponding author: nju.zhanghongjie@gmail.com}

\maketitle

\begin{center}
    \renewcommand{\arraystretch}{1.5}
    \vspace{1em}
    \begin{tabular}{rll}
        \github{} & \textbf{GitHub Repo} & \url{https://github.com/hmwang2002/CTRL-S} \\
        \huggingface{} & \textbf{SVG-Sophia Dataset} & \url{https://huggingface.co/datasets/InternSVG/SVG-Sophia} \\
    \end{tabular}
\end{center}

\section{Introduction}
\label{sec:intro}

\begin{figure}[htbp]
  \begin{center}
  \includegraphics[width=0.95\textwidth]{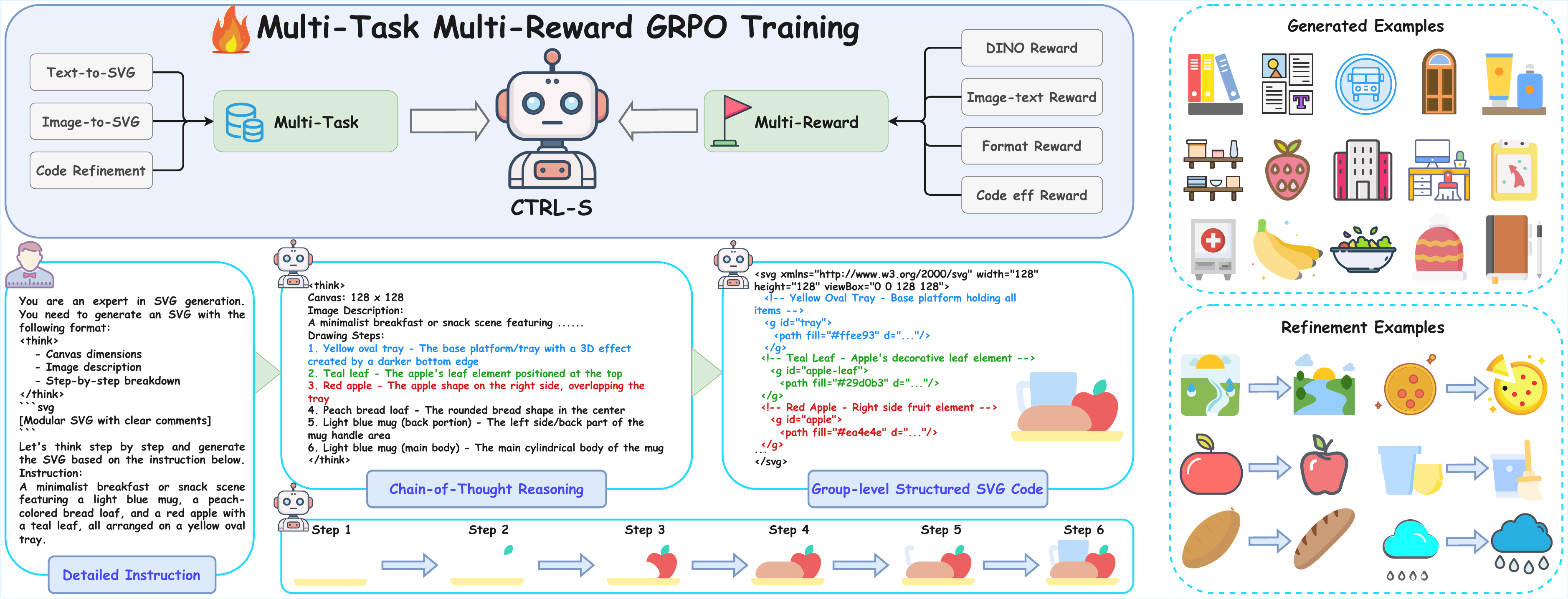}
  \caption{
    \textbf{Overview of \method.} \textbf{(Top Left)} The Multi-Task Multi-Reward GRPO training framework integrates diverse generation tasks (Text-to-SVG, Image-to-SVG, and Code Refinement) guided by multiple rewards. \textbf{(Bottom Left)} During inference, \method leverages chain-of-thought reasoning to plan step-by-step drawing operations before generating the final group-level structured SVG code, ensuring a clear one-to-one correspondence between the reasoning steps and the generated code groups. \textbf{(Right)} Examples of high-quality generated SVGs and successful code refinement processes.
  }
  \label{fig:overview}
  \end{center}
\end{figure}

Scalable Vector Graphics (SVG) is an XML-based vector format that represents 2D content using parameterized geometric primitives rather than pixel grids, offering compact storage, resolution independence, and fine-grained editability. Owing to its seamless integration with modern front-end systems and interactive frameworks, SVG has become a fundamental graphic medium in web design, user interface development, scientific visualization, and computer-aided design.

With the rapid development of vision-language models~\cite{gpt4o,meta2025llama4scout,meta2025llama4maverick,zhu2025internvl3,wang2025internvl3,bai2025qwen3}, recent research has begun to explore their application to high-quality SVG code generation~\cite{rodriguez2025starvector,xing2025empowering,yang2025omnisvg,wang2025internsvg}. By integrating vision encoders and SVG-specific tokens, these approaches significantly improve performance on Text-to-SVG and Image-to-SVG tasks. However, these approaches still suffer from limited generalization, frequently producing SVG programs with redundant paths. In addition, overly aggressive code compression during training degrades the readability and editability of the generated vector graphics. SVGen~\cite{wang2025svgen} and SVGThinker~\cite{chen2025svgthinker} introduce the chain-of-thought (CoT) reasoning into SVG generation by explicitly exposing intermediate reasoning steps to improve the quality of the generated SVG. However, they do not fully exploit the inherent grouping (\texttt{<g>}) structures in SVG code to organize components hierarchically, nor do they establish a clear alignment between reasoning steps and the corresponding grouped code segments, resulting in limited structural transparency and editability. While recent works like RLRF~\cite{rodriguez2025rendering} and Reason-SVG~\cite{xing2025reason} incorporate the GRPO algorithm~\cite{shao2024deepseekmath} to leverage visual reward signals during post-training reinforcement learning, they primarily optimize individual tasks in isolation and lack a unified framework for jointly training Text-to-SVG and Image-to-SVG generation.

To address these limitations, we propose \textbf{\method}, a unified framework tailored for Text-to-SVG, Image-to-SVG, and SVG code refinement tasks. As illustrated in Figure~\ref{fig:overview}, we integrate CoT reasoning into SVG generation to expose the model's planning processes. By leveraging the inherent grouping characteristics of SVG, we establish a step-wise alignment between the reasoning steps and the corresponding code groups. Furthermore, to break the isolation of prior works that exclusively focus on single-task optimization, we not only jointly train the Text-to-SVG and Image-to-SVG tasks but also introduce an SVG code refinement task. By endowing the model with self-diagnostic and error-correction capabilities, these three tasks mutually reinforce each other within a single unified model.

To facilitate this unified paradigm, we first construct \textbf{SVG-Sophia}, a high-quality dataset that encompasses CoT question-answering pairs across the three tasks. Comprising 131K SFT samples and 14.4K RL samples, SVG-Sophia provides a solid foundation for \method to excel in these diverse domains. In the RL post-training phase, we address the limitations of conventional SFT, which relies solely on token-level supervision and lacks visual feedback. We introduce a multi-task, multi-reward optimization framework based on the GRPO algorithm. Specifically, we design four complementary rewards: 
(1) a format reward to ensure structural validity and renderability,
(2) a DINO reward to encourage deep visual feature alignment between the rendered SVG and the reference image,
(3) an image–text similarity reward to promote semantic consistency between the generated SVG and the input instruction, and 
(4) a code efficiency reward to penalize unnecessarily verbose SVG outputs and improve inference efficiency. 
This multi-reward optimization not only enhances visual fidelity but also mitigates the repetitive code generation commonly observed in prior SVG-LLM models, achieving a balanced trade-off between reasoning efficiency and generation quality. Extensive experiments show that our multi-task, multi-reward RL algorithm yields significant gains over SFT. Joint multi-task training further improves performance and generalization compared to single-task optimization. Moreover, the introduction of CoT enhances generation success and visual quality for complex geometries, while transforming the implicit generation process into explicit, structured code blocks, substantially improving the readability and editability of the resulting SVGs. In summary, our contributions are as follows:

\begin{enumerate}
    \item We propose \textbf{\method}, a unified framework that integrates chain-of-thought reasoning and multi-task, multi-reward online RL for SVG code refinement, Text-to-SVG, and Image-to-SVG tasks. 
    \item We construct \textbf{SVG-Sophia}, a high-quality dataset providing explicit chain-of-thought supervision across three SVG tasks.
    \item Extensive experiments show that our multi-task, multi-reward RL framework achieves substantial performance gains over SFT baselines. \method achieves state-of-the-art performance in SVG generation, delivering higher visual quality, faster inference, and highly readable and editable code.
\end{enumerate}

\section{Related Work}
\label{sec:related_work}

\noindent\textbf{Optimization-based SVG Modeling.}
Optimization-based methods formulate SVG modeling as a parameter optimization problem rather than training a dedicated generative model. Early works such as DiffVG~\cite{li2020differentiable} and LIVE~\cite{ma2022towards} leverage differentiable rasterization to directly optimize Bézier control points and styling attributes by minimizing pixel-level reconstruction losses. To incorporate semantic supervision, CLIP-based approaches~\cite{frans2022clipdraw,schaldenbrand2022styleclipdraw,vinker2022clipasso,song2023clipvg,vinker2023clipascene} replace pixel losses with image-text similarity objectives, enabling text-conditioned SVG generation without training. More recently, Score Distillation Sampling (SDS)~\cite{poole2022dreamfusion} has been adopted to transfer diffusion priors into the vector graphics domain~\cite{jain2023vectorfusion,xing2023diffsketcher,zhang2024text,xing2024svgdreamer,xing2025svgdreamer++}. These methods optimize rendered SVGs through gradients derived from pretrained diffusion models, with later variants such as VPSD introducing particle-based distributional optimization to improve diversity and stability. Despite their strong visual fidelity, optimization-based approaches remain computationally intensive, instance-specific, and lack explicit hierarchical modeling of SVG structure, limiting scalability and downstream editability.

\noindent\textbf{Learning-based SVG Modeling.}~
Early learning-based methods represent SVG as sequences of geometric primitives and adopt task-specific generative architectures~\cite{ha2017neural,lopes2019learned,carlier2020deepsvg,reddy2021im2vec,ribeiro2020sketchformer,shen2021clipgen}. Sketch-RNN~\cite{ha2017neural} models drawings as sequential pen trajectories, SVG-VAE~\cite{lopes2019learned} introduces latent-variable modeling for vector synthesis, and DeepSVG~\cite{carlier2020deepsvg} employs hierarchical VAEs with Transformer decoders to capture global layouts and path-level details. With the emergence of large language models (LLMs) and vision-language models (VLMs), recent research has shifted toward semantically grounded SVG generation~\cite{wu2023iconshop,rodriguez2025starvector,xing2025empowering,chen2025svgbuilder,yang2025omnisvg,zou2024vgbench,li2025unisvg,wang2025svgen,chen2025svgenius,chen2025svgthinker,xing2025reason,rodriguez2025rendering,wang2025internsvg}. Methods like StarVector~\cite{rodriguez2025starvector}, LLM4SVG~\cite{xing2025empowering}, OmniSVG~\cite{yang2025omnisvg}, and InternSVG~\cite{wang2025internsvg} incorporate vision encoders and SVG-specific tokens to support Text-to-SVG and Image-to-SVG generation. Moreover, recent works such as SVGen~\cite{wang2025svgen} and SVGThinker~\cite{chen2025svgthinker} aim to introduce chain-of-thought reasoning into SVG generation by explicitly exposing intermediate reasoning steps, thereby improving performance. However, they fail to fully exploit the inherent grouping characteristics of SVG code to establish a one-to-one alignment between the intermediate planning steps and the generated code blocks.

\noindent\textbf{Reinforcement Learning for SVG Generation.}
Beyond standard supervised fine-tuning, applying reinforcement learning (RL) during the post-training stage has emerged as a promising frontier for SVG generation. Recent works such as RLRF~\cite{rodriguez2025rendering} and Reason-SVG~\cite{xing2025reason} adopt the GRPO algorithm~\cite{shao2024deepseekmath}, introducing visual reward signals to further enhance generative quality. However, these approaches remain confined to single-task optimization, failing to unify Text-to-SVG and Image-to-SVG generation under a shared paradigm. In contrast, our \method introduces a unified, multi-task RL optimization framework that jointly aligns Text-to-SVG, Image-to-SVG, and SVG code refinement within a single unified model.

\section{SVG-Sophia}
\label{sec:data}

We collect the original SVG files from the ColorSVG-100K~\cite{chen2025svgbuilder} dataset and leverage Claude-Sonnet-4.5~\cite{claude_4_5_sonnet} to annotate them into high-quality samples with explicit chain-of-thought reasoning and group-level structured SVG code. For Text-to-SVG generation, we construct 50K SFT samples and 5.5K RL samples. For Image-to-SVG generation, we similarly build 50K SFT samples and 5.5K RL samples, sharing the same underlying SVG programs as Text-to-SVG but differing in input modality. For SVG code refinement, we curate 31K SFT samples and 3.4K RL samples, along with a test set of 934 samples.

\subsection{Task Definition}
\label{subsec:data_definition}

Let $\mathcal{M}$ denote the MLLM and $I_{\text{text}}$ represent the user-provided textual instruction. For the \textbf{Text-to-SVG} generation task, the model is tasked with autoregressively generating a CoT planning sequence $O_{\text{think}}$, followed by the corresponding executable SVG code $O_{\text{svg}}$. This process is defined as:
\begin{equation}
    (O_{\text{think}}, O_{\text{svg}}) = \mathcal{M}(I_{\text{text}})
\end{equation}

Similarly, for the \textbf{Image-to-SVG} generation task, the model is additionally conditioned on a reference image $I_{\text{image}}$. The task is formulated as:

\begin{equation}
    (O_{\text{think}}, O_{\text{svg}}) = \mathcal{M}(I_{\text{text}}, I_{\text{image}})
\end{equation}

To empower the model with self-correction and optimization capabilities, we introduce the \textbf{SVG code refinement} task. In this setting, the model is provided with a textual instruction $I_{\text{text}}$, a reference image $I_{\text{image}}$, and a flawed SVG code draft $I_{\text{draft}}$ to be refined:

\begin{equation}
    (O_{\text{think}}, O_{\text{refined}}) = \mathcal{M}(I_{\text{text}}, I_{\text{image}}, I_{\text{draft}})
\end{equation}

\subsection{Data Annotation Pipeline}

The raw SVG files are initially collected from the ColorSVG-100K~\cite{chen2025svgbuilder} dataset and then normalized to a $128\times128$ viewBox. We employ Claude-Sonnet-4.5~\cite{claude_4_5_sonnet} to annotate detailed image captions from the rendered vector graphics. Subsequently, we prompt Claude-Sonnet-4.5 with both the generated caption and the raw SVG code, instructing it to refactor the original code into a highly structured format, enriched with descriptive comments and semantic group-level hierarchies, while also producing a step-by-step reasoning process that outlines its planning procedure. To ensure strict visual fidelity and eliminate failed refactoring attempts, we filter the refactored SVGs by retaining only those achieving an $\text{SSIM} \ge 0.95$ against their original renderings. To further ensure annotation quality, we engage 100 human annotators to review all annotated samples, manually correcting any captions that inaccurately describe the visual content or CoT reasoning steps that fail to correspond to the generated code groups. Finally, we use the generated image captions as user instructions and treat the CoT reasoning along with the reconstructed structured SVG code produced by Claude-Sonnet-4.5 as the ground-truth responses for Text-to-SVG and Image-to-SVG tasks.

For the SVG code refinement task, we first train a Qwen3-VL-8B model~\cite{bai2025qwen3} on the annotated Text-to-SVG and Image-to-SVG data, and use it to generate draft SVG programs on the training set. We then retain only moderately flawed samples ($0.30 \le \text{SSIM} \le 0.95$) against the ground truth. Claude-Sonnet-4.5 is then prompted with the defective and ground-truth images to produce discrepancy analysis and correction-oriented CoT reasoning. Rule-based filtering is further applied to remove invalid annotations, such as cases claiming complete consistency or providing irrelevant analysis. To mitigate potential annotator bias, 100 human annotators further review all refinement annotations, manually correcting cases where the identified defects or correction reasoning are inaccurate or task-irrelevant. For the test set, we select non-overlapping SVG programs from ColorSVG-100K and apply the same annotation pipeline. Defective drafts are generated using the SFT-trained Qwen3-VL-8B, as well as Claude-Sonnet-4.5, Gemini-3-Pro~\cite{gemini3}, GPT-5.2~\cite{gpt-5.2}, and Qwen3-VL-235B-A22B~\cite{bai2025qwen3}, to ensure a fair and unbiased evaluation.

\section{\method}
\label{sec:method}

\begin{figure}[t]
  \begin{center}
  \includegraphics[width=0.95\textwidth]{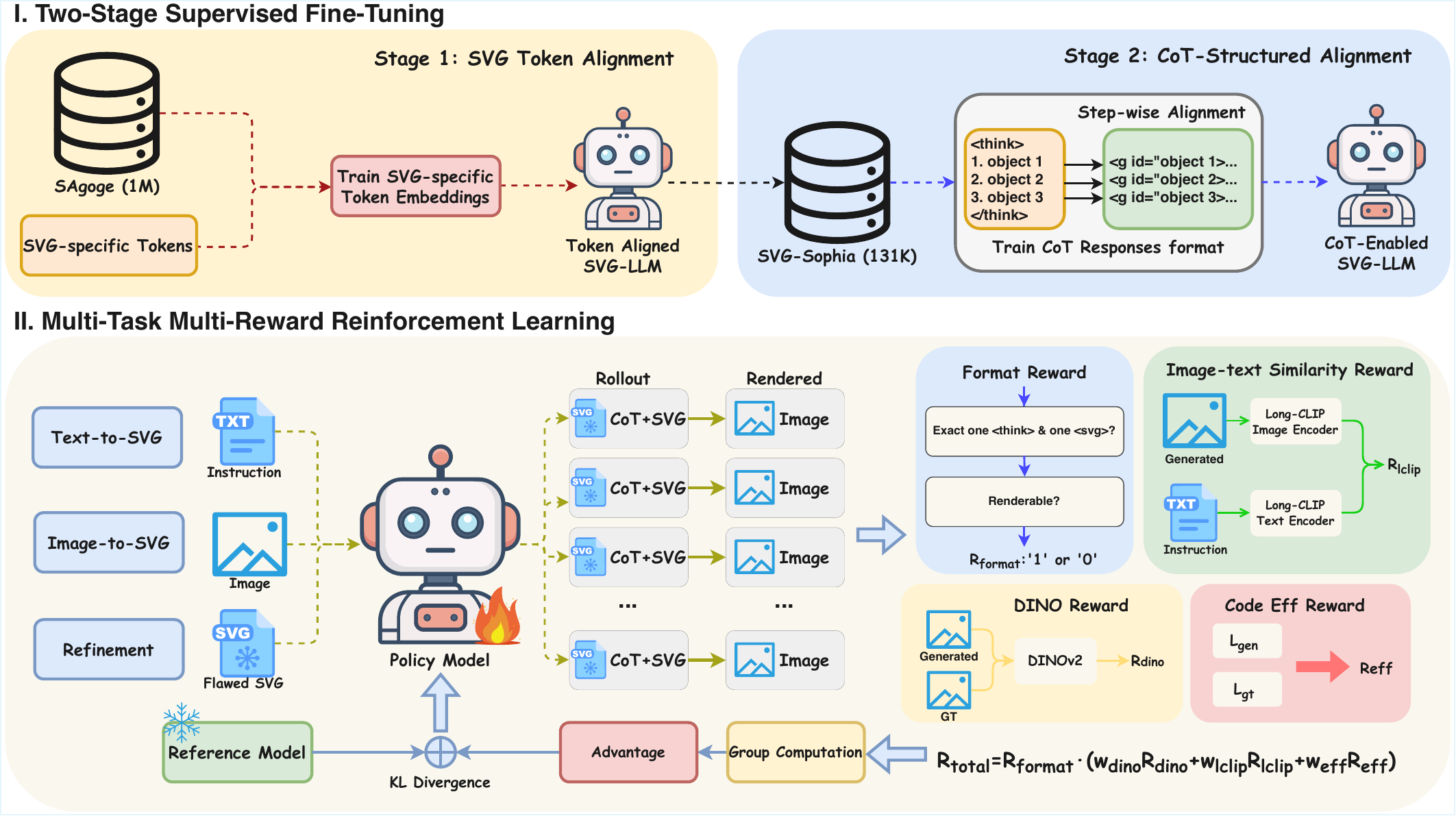}
  \caption{
    \textbf{The overall pipeline of \method.} \textbf{(1) Two-Stage SFT:} The model is first trained on 1M SAgoge samples to align SVG-specific tokens, and then fine-tuned on SVG-Sophia to learn CoT-structured responses with explicit step-wise planning. \textbf{(2) Multi-Task Multi-Reward RL:} We jointly optimize Text-to-SVG, Image-to-SVG, and SVG refinement tasks via a multi-reward mechanism, including Format Reward, DINO Reward, Image-text Similarity Reward, and Code Efficiency Reward, to improve structural validity, visual fidelity, semantic alignment, and concise code generation.
  }
  \label{fig:method}
  \end{center}
  \vspace{-1.5em}
\end{figure}

Figure~\ref{fig:method} illustrates the overall pipeline of \method. Our framework begins with a two-stage supervised fine-tuning to align SVG-specific tokens and establish step-wise chain-of-thought reasoning. Subsequently, a multi-task, multi-reward reinforcement learning phase jointly optimizes Text-to-SVG, Image-to-SVG, and code refinement tasks via comprehensive feedback signals.

\subsection{Preliminary}
\label{subsec:preliminary}
\noindent\textbf{Notation and Problem Formulation.}~
We formulate SVG generation as a unified multi-task sequence-to-sequence autoregressive generation problem. Let $\mathcal{M}$~(defined in Sec.~\ref{subsec:data_definition}) parameterized by $\theta$ denote our MLLM. Depending on the specific task, the model is conditioned on a varying set of inputs to generate a target sequence $y = (y_1, y_2, \dots, y_T)$, which consists of a chain-of-thought reasoning sequence $O_{\text{think}}$ followed by the executable SVG code ($O_{\text{svg}}$ or $O_{\text{refined}}$).
To unify our three core tasks, $c$ encapsulates varying inputs: $c = I_{\text{text}}$ for \textit{Text-to-SVG}, $c = (I_{\text{text}}, I_{\text{image}})$ for \textit{Image-to-SVG}, and $c = (I_{\text{text}}, I_{\text{image}}, I_{\text{draft}})$ for \textit{SVG Code Refinement}. Given the task-specific context $c$, the generation probability of the output sequence $y$ is factorized as:
\begin{equation}
\label{eq:definition}
    P(y|c) = \prod_{t=1}^{T} \pi_{\theta}(y_t | c, y_{<t}),
\end{equation}
where $y_{<t}$ represents the sequence of tokens generated prior to step $t$. The model, typically initialized after multi-task SFT, serves as our reference policy $\pi_{ref}$ for the reinforcement learning phase.

\noindent\textbf{Group Relative Policy Optimization (GRPO).}~
To efficiently optimize the MLLM across diverse tasks without the memory overhead of a parameterized value model, we employ GRPO~\cite{shao2024deepseekmath}. For a given context $c$, the current policy $\pi_{\theta_{old}}$ samples a group of $G$ diverse output trajectories $\mathcal{G} = \{y^{(1)}, \dots, y^{(G)}\}$. Each trajectory $y^{(i)}$ is evaluated by our multi-reward function to yield a score $r_i$. GRPO computes the relative advantage by normalizing these rewards within the group: $\hat{A}_i = (r_i - \mu_{\mathcal{G}}) / (\sigma_{\mathcal{G}} + \epsilon)$. The policy $\pi_{\theta}$ is then optimized by maximizing a clipped surrogate objective, augmented with a Kullback-Leibler (KL) divergence penalty to mitigate excessive deviation from $\pi_{ref}$:
\begin{equation}
\label{eq:grpo}
\mathcal{J}_{GRPO}(\theta) = \mathbb{E}_{c \sim \mathcal{D}, \mathcal{G} \sim \pi_{\theta_{old}}} \left[ \frac{1}{G} \sum_{i=1}^G \frac{1}{|y^{(i)}|} \sum_{t=1}^{|y^{(i)}|} \mathcal{L}_{clip}^{i,t}(\theta) - \beta \mathbb{D}_{KL}(\pi_\theta \| \pi_{ref}) \right],
\end{equation}
where the clipped likelihood ratio is defined as 
\begin{equation}
\label{eq:cliped_likelihood}
\mathcal{L}_{clip}^{i,t}(\theta) = \min \left( \rho_{i,t}(\theta) \hat{A}_i, \text{clip}(\rho_{i,t}(\theta), 1-\gamma, 1+\gamma) \hat{A}_i \right),
\end{equation}
and $\rho_{i,t}(\theta) = \frac{\pi_{\theta}(y_t^{(i)} | c, y_{<t}^{(i)})}{\pi_{\theta_{old}}(y_t^{(i)} | c, y_{<t}^{(i)})}$ is the probability ratio of generating the $t$-th token under the current versus the old policy.

\subsection{Two-Stage Supervised Fine-Tuning}
\label{subsec:method_sft}

To establish a robust initialization for the subsequent reinforcement learning phase, \method adopts the SVG-specific token design introduced in InternSVG~\cite{wang2025internsvg} (detailed in the Appendix) and undergoes a two-stage SFT process. 
In the first stage, we stabilize the embeddings of the SVG-specific tokens by sampling 1M training instances from the SAgoge dataset~\cite{wang2025internsvg}. 
Following this modality alignment, the second stage utilizes the SFT split of the SVG-Sophia dataset to train the model. This phase introduces a strict step-wise alignment, where each intermediate reasoning step in the CoT explicitly corresponds to a hierarchically organized, group-level (\texttt{<g>}) structural block in the resulting SVG, ensuring that the SVG generation process is both interpretable and logically transparent.

\subsection{Multi-Reward Design for Reinforcement Learning in \method}
\label{subsec:ctrl-s}
Following the SFT phase, we employ reinforcement learning to further align the model's generation with visual, semantic, and structural objectives.
To provide comprehensive guidance without relying on costly human annotations, we design a multi-reward framework comprising four complementary components.

\noindent\textbf{Format Reward ($R_{\text{format}}$)}~
To guarantee both structural compliance and execution validity, we introduce a binary format reward $R_{\text{format}}$. The reward yields 1 if the model's output strictly contains exactly one \texttt{<think>}...\texttt{</think>} reasoning block followed by a single SVG code block that can be rendered by CairoSVG successfully, and 0 otherwise.

\noindent\textbf{DINO Reward ($R_{\text{dino}}$)}~
A primary limitation of standard SFT is its inherent reliance on token-level textual supervision, which lacks the capacity to penalize \textit{global visual discrepancies}. For SVG-related tasks, explicit pixel-level feedback is crucial to enhance the overall visual fidelity of the generated graphics. To address this, we introduce $R_{\text{dino}}$. Specifically, the generated SVG code is first rasterized into an image $V_{\text{gen}}$. We then compute the feature similarity between this rendering and the ground-truth image $V_{\text{gt}}$ using a pre-trained DINOv2~\cite{dinov2} model, capturing deep, structural visual alignments. Formally, let $\mathcal{E}_{\text{DINO}}$ denote the DINOv2 feature extractor; the reward is formulated as the normalized cosine similarity between the two image embeddings:

\begin{equation}
\label{eq:dino_reward}
    R_{\text{dino}} = \frac{1}{2}(\cos(\mathcal{E}_{\text{DINO}}(V_{\text{gen}}), \mathcal{E}_{\text{DINO}}(V_{\text{gt}})) + 1).
\end{equation}

\noindent\textbf{Image-text Similarity Reward ($R_{\text{lclip}}$)}~
Beyond low-level visual fidelity~(Eq.~\ref{eq:dino_reward}), the generated SVG must semantically align with the user's high-level textual instruction $I_{\text{text}}$. Considering that the instructions in SVG-Sophia typically consist of several detailed descriptive sentences, the standard CLIP model~\cite{radford2021learning}, bounded by its strict 77-token input limit, often truncates crucial structural details and fails to adequately capture fine-grained semantics in long contexts.
To overcome this, we adopt Long-CLIP~\cite{zhang2024long} to compute the semantic alignment reward $R_{\text{lclip}}$. By leveraging the Long-CLIP image encoder $\mathcal{E}_{\text{L-img}}$ and text encoder $\mathcal{E}_{\text{L-text}}$, we project both the rendered image $V_{\text{gen}}$ and the instruction $I_{\text{text}}$ into a shared embedding space. The reward is computed as follows:

\begin{equation}
\label{eq:text_image_sim}
    R_{\text{lclip}} = \frac{\mathcal{E}_{\text{L-img}}(V_{\text{gen}})}{\|\mathcal{E}_{\text{L-img}}(V_{\text{gen}})\|_2} \cdot \frac{\mathcal{E}_{\text{L-text}}(I_{\text{text}})}{\|\mathcal{E}_{\text{L-text}}(I_{\text{text}})\|_2}.
\end{equation}

\noindent\textbf{Code Efficiency Reward ($R_{\text{eff}}$)}~
During the generation of SVG code, SFT models frequently suffer from a repetition problem, producing excessively long, redundant, and invalid code that significantly degrades inference speed. To mitigate this issue, we adapt a length-based penalty inspired by RLRF~\cite{rodriguez2025rendering}. Specifically, let $L_{\text{gt}}$ and $L_{\text{gen}}$ denote ground-truth and generated SVG code lengths, the code efficiency reward $R_{\text{eff}}$ is formulated as follows:

\begin{equation}
\label{eq:code_effiency}
    R_{\text{eff}} = 1 - (\frac{1}{L_{\text{gt}}}\max(0, L_{\text{gen}} - \frac{L_{\text{gt}}}{2}))^2.
\end{equation}

\noindent\textbf{Total Reward ($R_{\text{total}}$)}~
Finally, we aggregate the visual~(Eq.~\ref{eq:dino_reward}), semantic~(Eq.~\ref{eq:text_image_sim}), and efficiency objectives~(Eq.~\ref{eq:code_effiency}) into a unified multi-reward formulation. Crucially, the binary format reward $R_{\text{format}}$ acts as a multiplicative gating factor, ensuring that unrenderable or structurally malformed outputs receive a total reward of zero, preventing degenerate policy updates. The final reward is defined as:

\begin{equation}
\label{eq:overall_objective}
    R_{\text{total}} = R_{\text{format}} \cdot \left(
w_{\text{dino}} R_{\text{dino}}
+
w_{\text{lclip}} R_{\text{lclip}}
+
w_{\text{eff}} R_{\text{eff}}
\right).
\end{equation}
Empirically, we set the trade-off weights as $w_{\text{dino}} : w_{\text{lclip}} : w_{\text{eff}} = 2 : 1 : 1$.
\section{Experiments}
\label{sec:exp}

\subsection{Experimental Setup}

Building upon Qwen3-VL-8B-Instruct, \method initially undergoes a two-stage SFT process, as detailed in Sec.~\ref{subsec:method_sft}. We set the learning rate to 1e-4 in the first stage and decrease it to 5e-5 in the second stage. The training is performed on 48 H200 GPUs with a global batch size of 96. In the RL stage, we optimize the model using the GRPO algorithm implemented via the \texttt{verl} framework. The RL training is performed on 32 GPUs with a global batch size of 128 and a learning rate of 1e-5. During the rollout phase, we sample 16 responses per prompt. The model is trained for 2 epochs, and the entire RL training process takes approximately 12 hours.

\subsection{Quantitative Evaluations}


\begin{table}[t]
    \centering
    \small
    \setlength{\tabcolsep}{3pt}
    \renewcommand{\arraystretch}{1.2}
        \caption{SVG generation results on SArena-Icon. SR denotes Success Rate.
        \method~(SFT) denotes the model after two-stage SFT, \method~(SFT+RL) denotes the full model after RL post-training.}
    \label{tab:sarena-icon-gen}
    \resizebox{\textwidth}{!}{
    \begin{tabular}{c|c|ccccc|ccccc}
        \mytoprule
        \multirow{2}{*}{\textbf{Model}}
            & \multirow{2}{*}{\textbf{Data}}
            & \multicolumn{5}{c|}{\textbf{Text-to-SVG}} 
            & \multicolumn{5}{c}{\textbf{Image-to-SVG}} \\
        & & \textbf{FID $\downarrow$} 
        & \textbf{CLIP-T2I $\uparrow$} & \textbf{CLIP-I2I $\uparrow$}
        & \textbf{SR} & \textbf{Tokens} 
        & \textbf{DINO $\uparrow$} & \textbf{SSIM $\uparrow$} 
        & \textbf{LPIPS $\downarrow$}
        & \textbf{SR} & \textbf{Tokens} \\
        \mymidrule
        \rowcolor{oursblue} \multicolumn{12}{c}{\textbf{Traditional SVG Methods}} \\
        IconShop & --
            & 32.288 & 20.894 & 70.922 & 86.51\% & 1.6k
            & -- & -- & -- & -- & -- \\
        VectorFusion & --
            & 16.594 & 22.992 & 70.308 & 100.0\% & 33k
            & -- & -- & -- & -- & -- \\
        SVGDreamer & --
            & 26.612 & 20.329 & 69.975 & 100.0\% & 132k
            & -- & -- & -- & -- & -- \\
        DiffVG & --
            & -- & -- & -- & -- & --
            & 0.869 & 0.927 & 0.097 & 100.0\% & 17k \\
        LIVE & --
            & -- & -- & -- & -- & --
            & 0.973 & 0.986 & 0.024 & 100.0\% & 18k \\
        VTracer & --
            & -- & -- & -- & -- & --
            & 0.966 & 0.875 & 0.054 & 100.0\% & 4.4k \\
        \mymidrule
        \rowcolor{oursblue} \multicolumn{12}{c}{\textbf{Vision-Language Models}} \\
        Llama 4 Scout & --
            & 17.908 & 22.849 & 73.563 & 96.03\% & 256
            & 0.844 & 0.582 & 0.346 & 97.61\% & 246 \\
        Llama 4 Maverick & --
            & 14.931 & 23.570 & 75.816 & 98.12\% & 265
            & 0.863 & 0.596 & 0.329 & 97.75\% & 255 \\
        Qwen3-VL-235B-A22B & --
            & 18.404 & 24.024 & 76.289 & 94.96\% & 326
            & 0.927 & 0.718 & 0.210 & 95.28\% & 679 \\
        GPT-5.2 & --
            & 14.965 & 25.457 & 82.701 & 97.02\% & 623
            & 0.936 & 0.653 & 0.269 & 98.87\% & 497 \\
        Claude-Sonnet-4.5 & --
            & 16.451 & 25.322 & 81.038 & 99.63\% & 320
            & 0.910 & 0.677 & 0.253 & 97.95\% & 705 \\
        Gemini-3-Pro & --
            & 13.203 & 25.731 & 84.157 & 96.91\% & 397
            & 0.940 & 0.723 & 0.198 & 97.14\% & 769 \\
        \mymidrule
        \rowcolor{oursblue} \multicolumn{12}{c}{\textbf{SVG-LLMs}} \\
        StarVector-8B & 2.2M
            & -- & -- & -- & -- & --
            & 0.871 & 0.623 & 0.206 & 72.51\% & 951 \\
        LLM4SVG-7B & 580K
            & 21.939 & 19.458 & 70.726 & 90.95\% & 705
            & 0.748 & 0.472 & 0.409 & 95.53\% & 485 \\
        OmniSVG-3B & 2M
            & 28.292 & 21.679 & 74.831 & 99.68\% & 1.8k
            & 0.894 & 0.756 & 0.186 & 99.97\% & 2.4k \\
        InternSVG-8B & 16M
            & 8.715 & 23.916 & 80.911 & 97.24\% & 1.0k
            & 0.949 & 0.811 & 0.127 & 94.45\% & 1.3k \\
        \mymidrule
        \rowcolor{oursblue} \multicolumn{12}{c}{\textbf{Ours}} \\
        Qwen3-VL-8B & --
            & 16.875 & 22.578 & 73.187 & 99.55\% & 374
            & 0.796 & 0.500 & 0.354 & 74.02\% & 474 \\
        \method~(SFT) & 131K
            & 18.154 & 23.655 & 77.959 & 92.02\% & 1.1k
            & 0.903 & 0.717 & 0.187 & 90.12\% & 1.4k \\
        \rowcolor{oursgreen}
        \textbf{\method~(SFT+RL)} & 145K
            & \textbf{11.584} & \textbf{25.944} & \textbf{82.291} & \textbf{99.85\%} & 346
            & \textbf{0.980} & \textbf{0.835} & \textbf{0.098} & \textbf{99.93\%} & 512 \\
        \multicolumn{2}{c|}{\textsl{$\Delta$ Improvement}}
        & \impv{6.570} & \impv{2.289} & \impv{4.332} & {} & \multicolumn{1}{c|}{}
        & \impv{0.077} & \impv{0.118} & \impv{0.089} & {} & \multicolumn{1}{c}{} \\
        \mybottomrule
    \end{tabular}
    }
\end{table}

As shown in Table~\ref{tab:sarena-icon-gen}, \method achieves leading performance on both the Text-to-SVG and Image-to-SVG tasks on the SArena-Icon~\cite{wang2025internsvg} benchmark. For Text-to-SVG, \method attains the highest CLIP-T2I score of 25.944, demonstrating the outstanding semantic understanding and text-to-image alignment capabilities of the model. For Image-to-SVG, our model obtains the best results across the DINO, SSIM, and LPIPS metrics compared to mainstream general VLMs and SVG-LLMs. This proves that \method exhibits extremely high visual fidelity in accurately reconstructing geometric shapes and colors. Furthermore, compared to our SFT baseline, \method (SFT + RL) not only gains significant enhancements across all visual metrics but also substantially increases the success rate and greatly reduces the number of generated tokens. This validates the effectiveness of our proposed multi-task, multi-reward RL training framework. The framework successfully enhances visual fidelity and semantic alignment while simultaneously empowering the model with stronger generalization capabilities, robustness, and superior code generation efficiency.

As shown in Table~\ref{tab:sarena-icon-refine}, on the SVG-Sophia Code Refinement Benchmark, \method achieves the best performance across all evaluation metrics compared with state-of-the-art proprietary models, including GPT-5.2, Claude-Sonnet-4.5, and Gemini-3-Pro. These results highlight the superiority of \method in structural understanding and fine-grained code editing. Specifically, starting from imperfect SVG programs, our method accurately locates defective components and applies targeted corrections, achieving more precise structural refinement while preserving semantic consistency. This indicates that the model not only possesses a strong understanding of the visual information encoded in SVG programs but also demonstrates stable, iterative code-optimization capability. Furthermore, compared with the SFT baseline, the RL-enhanced model achieves a substantial improvement in task success rate while significantly reducing the number of generated tokens. This suggests better generalization and a reduced tendency to produce redundant code, leading to more efficient and controllable refinement.

\begin{table}[t]
    \centering
    \small
    \setlength{\tabcolsep}{4pt}
    \renewcommand{\arraystretch}{1.2}
    \captionsetup{justification=centering}
    \caption{SVG refinement results on SVG-Sophia Code Refinement Benchmark.}
    \label{tab:sarena-icon-refine}
    \begin{tabular}{c|ccccc}
        \mytoprule
        \textbf{Model} 
            & \textbf{DINO $\uparrow$} & \textbf{SSIM $\uparrow$} 
            & \textbf{LPIPS $\downarrow$}
            & \textbf{SR} & \textbf{Tokens} \\
        \mymidrule
        \rowcolor{oursblue} \multicolumn{6}{c}{\textbf{Vision-Language Models}} \\
        Llama 4 Scout
            & 0.840 & 0.634 & 0.383 & 97.00\% & 788 \\
        Llama 4 Maverick
            & 0.845 & 0.615 & 0.377 & 94.75\% & 616 \\
        Qwen3-VL-235B-A22B
            & 0.809 & 0.515 & 0.403 & 77.84\% & 840 \\
        GPT-5.2
            & 0.911 & 0.640 & 0.342 & 99.26\% & 975 \\
        Claude-Sonnet-4.5
            & 0.796 & 0.579 & 0.401 & 84.58\% & 790 \\
        Gemini-3-Pro
            & 0.883 & 0.593 & 0.284 & 81.37\% & 992 \\
        \mymidrule
        \rowcolor{oursblue} \multicolumn{6}{c}{\textbf{Ours}} \\
        Qwen3-VL-8B
            & 0.796 & 0.501 & 0.410 & 76.77\% & 980 \\
        \method (SFT)
            & 0.888 & 0.665 & 0.236 & 84.37\% & 2.9k \\
        \rowcolor{oursgreen}
        \textbf{\method~(SFT+RL)}
            & \textbf{0.951} & \textbf{0.765} & \textbf{0.180} & \textbf{99.79\%} & 866 \\
        \multicolumn{1}{c|}{\textsl{$\Delta$ Improvement}}
            & \impv{0.067} & \impv{0.126} & \impv{0.104} & {} & \multicolumn{1}{c}{} \\
        \mybottomrule
    \end{tabular}
\end{table}

\subsection{Qualitative Evaluations}

\begin{figure}[htbp]
    \centering
    
    \begin{subfigure}{1.0\textwidth}
        \centering
        \includegraphics[width=\textwidth]{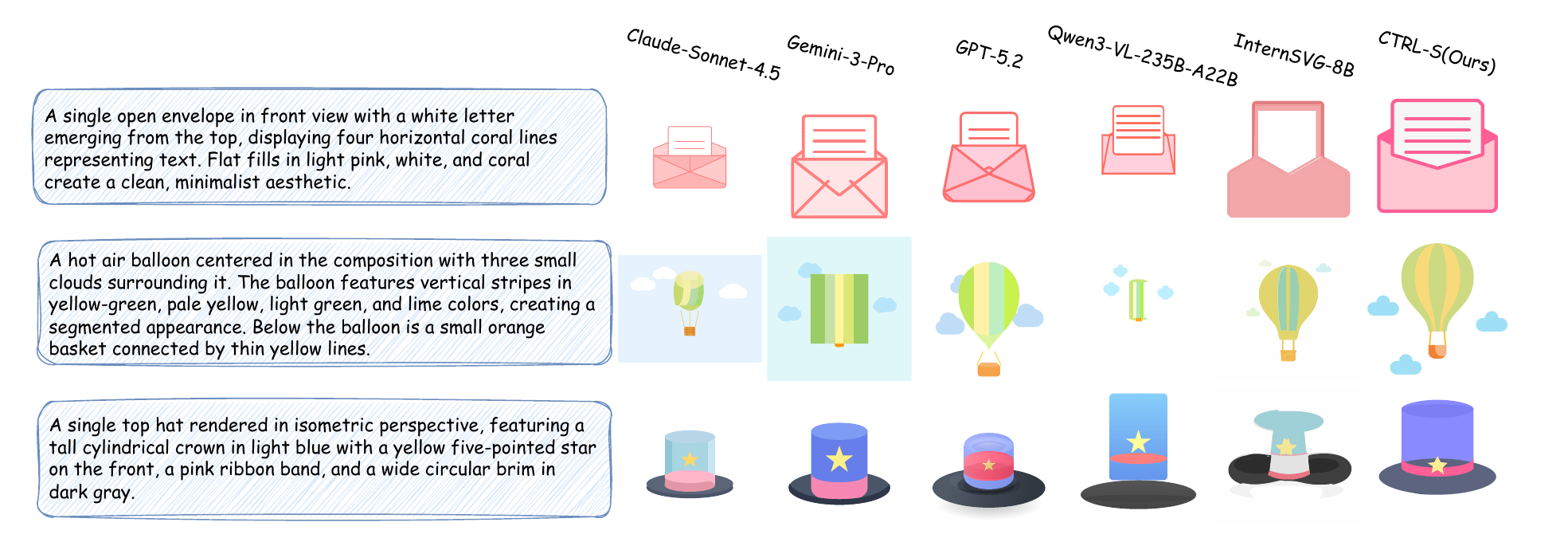}
        \captionsetup{justification=centering}
        \caption{Qualitative comparison on Text-to-SVG generation.}
        \label{fig:qual-t2s}
    \end{subfigure}
    
    \vspace{3mm}
    
    \begin{subfigure}{1.0\textwidth}
        \centering
        \includegraphics[width=\textwidth]{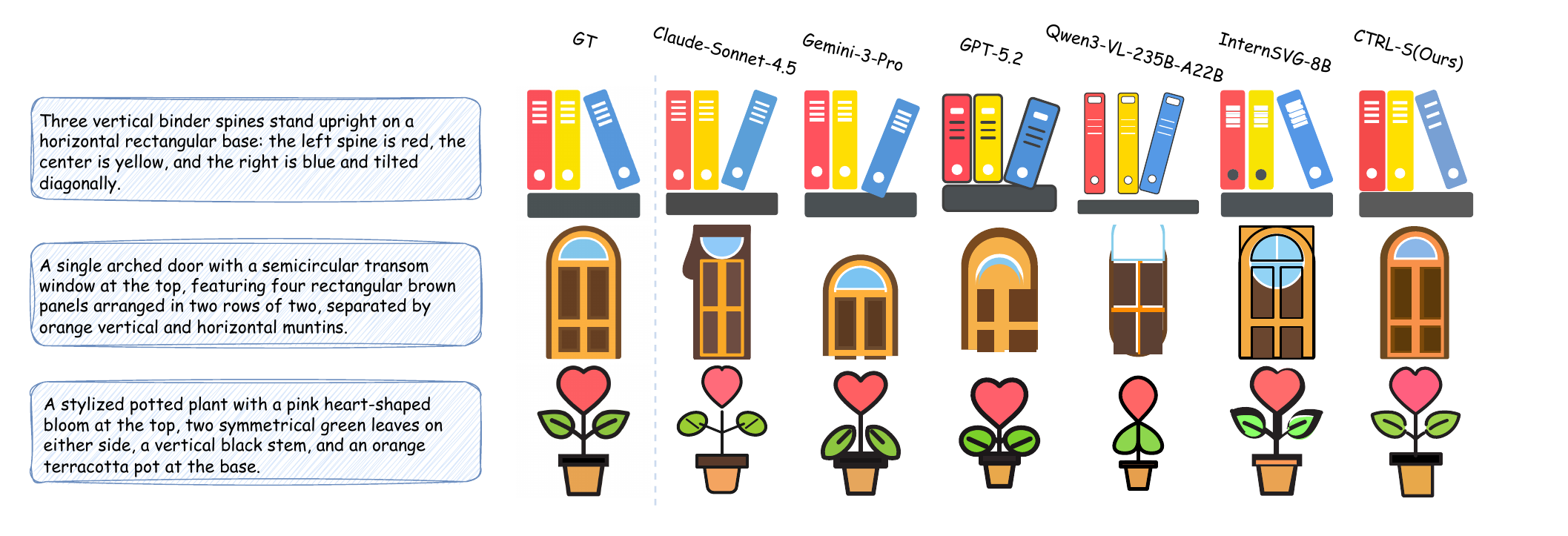}
        \captionsetup{justification=centering}
        \caption{Qualitative comparison on Image-to-SVG generation.}
        \label{fig:qual-i2s}
    \end{subfigure}
    
    \vspace{3mm}
    
    \begin{subfigure}{1.0\textwidth}
        \centering
        \includegraphics[width=\textwidth]{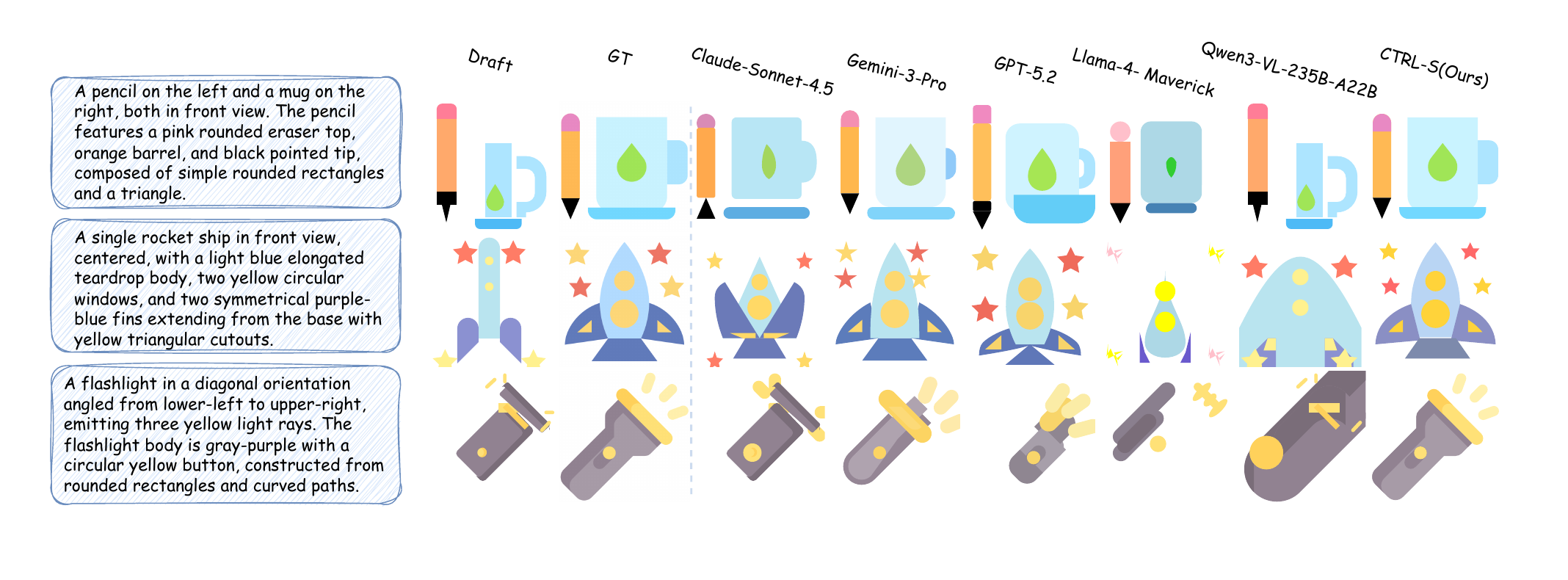}
        \captionsetup{justification=centering}
        \caption{Qualitative comparison on SVG code refinement.}
        \label{fig:qual-refine}
    \end{subfigure}
    
    \caption{Qualitative comparisons of SVG generation and code refinement between baselines and \method.}
    \label{fig:qualitative}
\end{figure}

\begin{figure}[t]
    \centering
    \includegraphics[width=\textwidth]{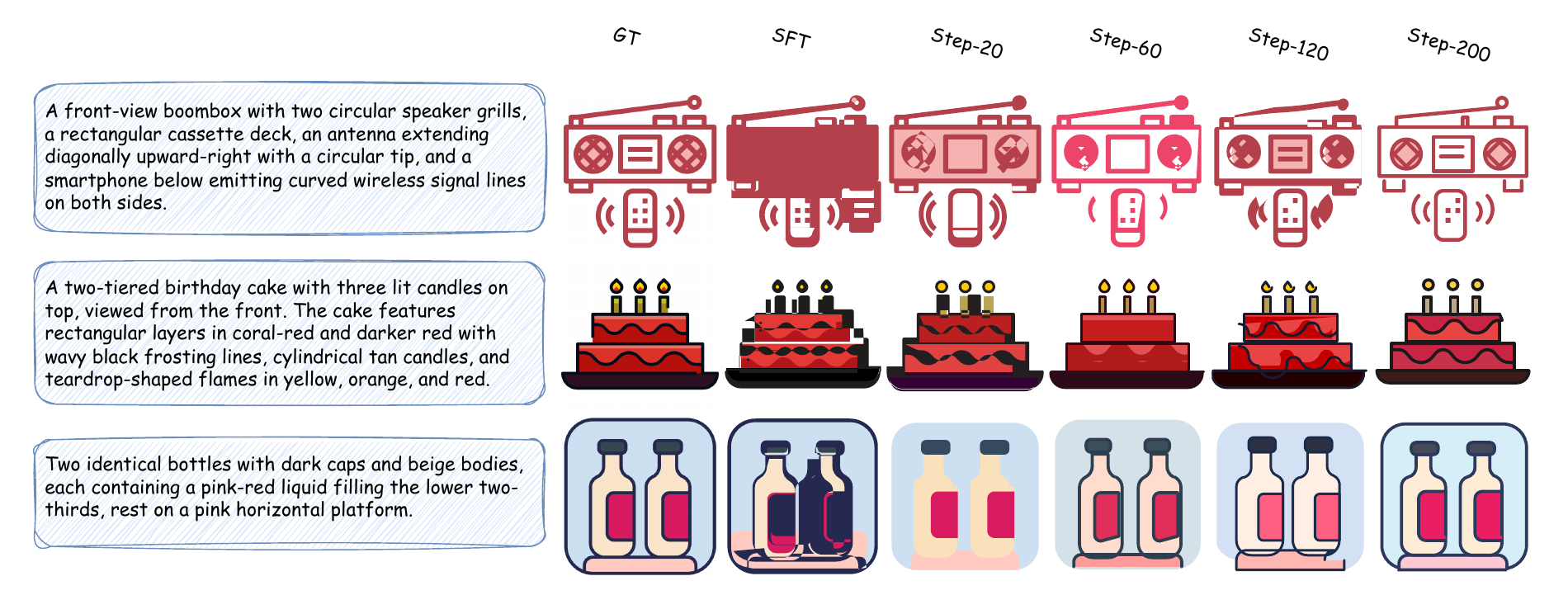}
    \captionsetup{justification=centering}
    \caption{Qualitative visualization of SVG generation quality across RL training steps.}
    \label{fig:rl_steps}
\end{figure}

As shown in Figure~\ref{fig:qualitative}, we present qualitative comparisons between \method and representative baselines across three tasks.
For Text-to-SVG and Image-to-SVG tasks (Figures~\ref{fig:qual-t2s} and~\ref{fig:qual-i2s}), \method consistently generates SVG with accurate structural layouts and fine-grained visual details. For Text-to-SVG, \method accurately renders complex multi-element scenes, such as a striped hot air balloon with surrounding clouds and a suspended basket. In contrast, competing methods frequently distort the global structure, omit essential components, or fail to reproduce the correct color patterns. For Image-to-SVG, \method faithfully reconstructs reference images while preserving color attributes and spatial arrangements. In contrast, general VLMs often produce geometries that are structurally inconsistent and physically implausible, while SVG-LLMs such as InternSVG-8B exhibit noticeable deviations in stroke thickness and shape boundaries. 
For SVG code refinement (Figure~\ref{fig:qual-refine}), \method demonstrates strong self-correction capability by accurately identifying structural defects in the draft SVG and applying targeted modifications. In contrast, general VLMs often fail to precisely localize defective components, resulting in incomplete or globally inconsistent corrections. Starting from imperfect drafts, \method successfully recovers missing components and corrects spatial misalignments, producing refined outputs that closely
match the ground truth. In Figure~\ref{fig:rl_steps}, we further illustrate the progressive improvement of \method throughout RL training. As the number of training steps increases, the generated SVGs exhibit increasingly accurate structural layouts, more faithful color reproduction, and richer fine-grained details, demonstrating the effectiveness of our multi-reward optimization in iteratively refining the model's generation capability.

\subsection{Ablation Studies}

\begin{table}[t]
    \centering
    \small
    \setlength{\tabcolsep}{4pt}
    \renewcommand{\arraystretch}{1.2}
    \captionsetup{justification=centering}
    \caption{Ablation studies on CoT, reward signals, reward ratios, and multi-task training.}
    \label{tab:ablation}

    \begin{subtable}{\textwidth}
        \centering
        \caption{Ablation on Chain-of-Thought (CoT).}
        \label{tab:ablation-cot}
        \vspace{-2mm}
        \resizebox{\textwidth}{!}{
        \begin{tabular}{l|ccccc|ccccc}
            \mytoprule
            \multirow{2}{*}{\textbf{Model}}
                & \multicolumn{5}{c|}{\textbf{Text-to-SVG}}
                & \multicolumn{5}{c}{\textbf{Image-to-SVG}} \\
            & \textbf{FID $\downarrow$}
            & \textbf{CLIP-T2I $\uparrow$} & \textbf{CLIP-I2I $\uparrow$}
            & \textbf{SR} & \textbf{Tokens}
            & \textbf{DINO $\uparrow$} & \textbf{SSIM $\uparrow$}
            & \textbf{LPIPS $\downarrow$}
            & \textbf{SR} & \textbf{Tokens} \\
            \mymidrule
            w/o CoT (SFT)
                & 22.960 & 23.002 & 76.567 & 85.75\% & 1.0k
                & 0.887 & 0.663 & 0.204 & 81.96\% & 1.3k \\
            \rowcolor{oursgreen}
            w/ CoT (SFT)
                & 18.154 & 23.655 & 77.959 & 92.02\% & 1.1k
                & 0.903 & 0.717 & 0.187 & 90.12\% & 1.4k \\
            \mymidrule
            w/o CoT (SFT+RL)
                & 12.161 & 24.278 & 80.604 & 99.47\% & 274
                & 0.961 & 0.813 & 0.125 & 99.53\% & 466 \\
            \rowcolor{oursgreen}
            w/ CoT (SFT+RL)
                & \textbf{11.584} & \textbf{25.944} & \textbf{82.291} & \textbf{99.85\%} & 346
                & \textbf{0.980} & \textbf{0.835} & \textbf{0.098} & \textbf{99.93\%} & 512 \\
            \mybottomrule
        \end{tabular}
        }
    \end{subtable}
    
    \vspace{4mm}

    \begin{subtable}{0.48\textwidth}
        \centering
        \caption{Ablation on multi-reward functions.}
        \label{tab:ablation-reward}
        \vspace{-2mm}
        \resizebox{\textwidth}{!}{
        \begin{tabular}{l|ccccc}
            \mytoprule
            \textbf{Reward}
                & \textbf{FID $\downarrow$} & \textbf{CLIP-T2I $\uparrow$}
                & \textbf{CLIP-I2I $\uparrow$} & \textbf{Tokens}
                & \textbf{Time (s/svg) } \\
            \mymidrule
            $R_{\text{format}} \cdot R_{\text{dino}}$
                & 12.443 & 24.573 & 80.897 & 606 & 6.151 \\
            $R_{\text{format}} \cdot (R_{\text{dino}} + R_{\text{lclip}})$
                & 11.831 & 25.444 & 82.070 & 701 & 7.121 \\
            \rowcolor{oursgreen}
            Full $R_{\text{total}}$ (Ours)
                & \textbf{11.584} & \textbf{25.944} & \textbf{82.291} & \textbf{346} & \textbf{4.439} \\
            \mybottomrule
        \end{tabular}
        }
    \end{subtable}
    \hfill
    \begin{subtable}{0.5\textwidth}
        \centering
        \caption{Ablation on reward ratio ($w_{\text{dino}}$:$w_{\text{lclip}}$:$w_{\text{eff}}$).}
        \label{tab:ablation-ratio}
        \vspace{-2mm}
        \resizebox{\textwidth}{!}{
        \begin{tabular}{c|cccc|cccc}
            \mytoprule
            \multirow{2}{*}{\textbf{Ratio}}
                & \multicolumn{4}{c|}{\textbf{Text-to-SVG}}
                & \multicolumn{4}{c}{\textbf{Image-to-SVG}} \\
            & \textbf{FID $\downarrow$} & \textbf{CLIP-T2I $\uparrow$}
            & \textbf{CLIP-I2I $\uparrow$} & \textbf{Tokens}
            & \textbf{DINO $\uparrow$} & \textbf{SSIM $\uparrow$}
            & \textbf{LPIPS $\downarrow$} & \textbf{Tokens} \\
            \mymidrule
            1:1:1
                & 12.575 & 24.623 & 80.849 & 333
                & 0.962 & 0.818 & 0.120 & 532 \\
            5:3:1
                & 11.642 & 24.653 & 81.336 & 428
                & 0.967 & 0.830 & 0.109 & 649 \\
            \rowcolor{oursgreen}
            2:1:1 (Ours)
                & \textbf{11.584} & \textbf{25.944} & \textbf{82.291} & 346
                & \textbf{0.980} & \textbf{0.835} & \textbf{0.098} & \textbf{512} \\
            \mybottomrule
        \end{tabular}
        }
    \end{subtable}

    \vspace{4mm}

    \begin{subtable}{\textwidth}
        \centering
        \captionsetup{justification=raggedright}
        \caption{Ablation on multi-task training across Text-to-SVG, Image-to-SVG, and Refinement tasks. T2S denotes Text-to-SVG, I2S denotes Image-to-SVG, and Refine denotes SVG code refinement.}
        \label{tab:ablation-multitask}
        \vspace{-2mm}
        \resizebox{\textwidth}{!}{
        \begin{tabular}{l|ccc|ccc|ccc}
            \mytoprule
            \multirow{2}{*}{\textbf{Training}}
                & \multicolumn{3}{c|}{\textbf{Text-to-SVG}}
                & \multicolumn{3}{c|}{\textbf{Image-to-SVG}}
                & \multicolumn{3}{c}{\textbf{Refinement}} \\
            & \textbf{FID $\downarrow$}
            & \textbf{CLIP-T2I $\uparrow$} & \textbf{CLIP-I2I $\uparrow$}
            & \textbf{DINO $\uparrow$} & \textbf{SSIM $\uparrow$}
            & \textbf{LPIPS $\downarrow$}
            & \textbf{DINO $\uparrow$} & \textbf{SSIM $\uparrow$}
            & \textbf{LPIPS $\downarrow$} \\
            \mymidrule
            T2S Only
                & 12.321 & 24.431 & 80.651
                & -- & -- & --
                & -- & -- & -- \\
            I2S Only
                & -- & -- & --
                & 0.959 & 0.824 & 0.119
                & -- & -- & -- \\
            Refine Only
                & -- & -- & --
                & -- & -- & --
                & 0.940 & 0.753 & 0.187 \\
            T2S + I2S 
                & 11.637 & 24.804 & 81.433 
                & 0.966  & 0.824  & 0.114 
                & -- & -- & -- \\
            \rowcolor{oursgreen}
            Multi-Task (Ours)
                & \textbf{11.584} & \textbf{25.944} & \textbf{82.291}
                & \textbf{0.980} & \textbf{0.835} & \textbf{0.098}
                & \textbf{0.951} & \textbf{0.765} & \textbf{0.180} \\
            \mybottomrule
        \end{tabular}
        }
    \end{subtable}
\end{table}

\noindent\textbf{Benefits of Chain-of-Thought Reasoning.}~
To evaluate the impact of CoT reasoning on model performance, we construct a non-CoT dataset by directly using the original SVG codes from ColorSVG without reconstruction by Claude-Sonnet-4.5. Based on this dataset, we train baseline SFT and RL models without CoT capabilities. As shown in Table~\ref{tab:ablation-cot}, for the SFT models, introducing CoT reasoning substantially improves the task success rate. This indicates that incorporating explicit reasoning enhances generation robustness, enabling the model to produce more renderable and syntactically valid SVG programs. Moreover, for both SFT and RL settings, the CoT-enabled models consistently outperform their non-CoT counterparts across all visual quality metrics. These results suggest that CoT reasoning plays a critical role not only in improving generation stability but also in enhancing visual fidelity and semantic alignment.

\noindent\textbf{Ablation on Multi-Reward Functions.}~
To systematically investigate the specific contributions of individual reward functions, we conduct an ablation study on the reward mechanism, sequentially comparing the performance of models trained with: (1) only the Format and DINO Rewards, (2) the addition of the Image-Text Similarity Reward, and (3) the full composite reward. As demonstrated in Table~\ref{tab:ablation-reward}, for the Text-to-SVG task, incorporating the Image-Text Similarity Reward significantly enhances the model's semantic comprehension and text-image alignment capabilities, improving the CLIP-T2I score from 24.573 to 25.444 and the CLIP-I2I score from 80.897 to 82.070. More importantly, with the further integration of the Code Efficiency Reward, the model effectively mitigates the issue of code redundancy. It not only maintains superior image generation quality but also significantly accelerates the per-sample inference speed, reducing the generated tokens from 701 to 346 and the inference time from 7.121 to 4.439 seconds per SVG. Ultimately, this achieves an optimal balance between visual fidelity and inference efficiency.

\noindent\textbf{Ablation on Reward Ratio.}~
As demonstrated in Table~\ref{tab:ablation-ratio}, we empirically evaluate different weight ratios among $w_{\text{dino}}$, $w_{\text{lclip}}$, and $w_{\text{eff}}$. Our findings indicate that the 2:1:1 configuration achieves the most effective balance among visual quality, semantic alignment, and code efficiency. Specifically, increasing $w_{\text{dino}}$ favors structural consistency but leads to longer outputs, while reducing it harms geometric fidelity. We therefore adopt 2:1:1 in all experiments.

\noindent\textbf{Effects of Multi-task Training.}~
We analyze the impact of joint training across Text-to-SVG, Image-to-SVG, and SVG code refinement tasks. As shown in Table~\ref{tab:ablation-multitask}, compared with single-task training, multi-task learning significantly improves the overall quality of generated SVG. Jointly training Text-to-SVG and Image-to-SVG enhances cross-modal semantic alignment, leading to better consistency between textual instructions, input images, and generated SVG. Furthermore, incorporating the SVG code refinement task brings substantial gains to the Image-to-SVG setting. This suggests that learning to analyze rendered images from imperfect SVG and perform targeted SVG correction strengthens the model’s ability to reconstruct vector graphics from images. Overall, these results indicate that the three tasks provide complementary supervision signals, and their joint optimization leads to more robust and generalized SVG modeling.

\section{Conclusion}

In this work, we present \method, a unified framework that introduces chain-of-thought (CoT) reasoning into Text-to-SVG, Image-to-SVG, and SVG code refinement tasks. By leveraging the hierarchical grouping structure inherent in SVG, we explicitly align planning steps with code groups, thereby improving the structural consistency, readability, and editability of generated SVG. Building upon this design, we propose a multi-task, multi-reward GRPO training paradigm that jointly optimizes complementary tasks under diverse reward signals. Through cross-task collaboration and multi-perspective reward guidance, our approach significantly enhances visual fidelity, semantic alignment, and generation stability. In addition, we construct the SVG-Sophia dataset, a high-quality SVG corpus augmented with explicit CoT question–answer pairs, providing systematic training resources for structured SVG generation and code refinement. Extensive experimental results demonstrate that \method consistently achieves state-of-the-art performance across multiple tasks, validating the effectiveness of structured reasoning and multi-task, multi-reward optimization for vector graphic modeling. Overall, this work establishes a new training paradigm for reliable reasoning in SVG-LLMs and lays the foundation for future research in complex vector graphic generation and editing.
\section*{Acknowledgements}

This work was supported by the Shanghai Artificial Intelligence Laboratory and the Shanghai Committee of Science and Technology (No. 22YF1461500).

\clearpage
\bibliographystyle{plain}
\bibliography{refs}


\clearpage
\appendix
\section{Appendix}

\subsection{Design of SVG-specific tokens}

In this section, we summarize the design of SVG-specific tokens adopted in \method. As mentioned in the main paper, all SVGs are first mapped to a normalized $128 \times 128$ coordinate space. This unified parameterization reduces unnecessary variation in the absolute coordinate scale, simplifies geometric modeling, and improves consistency across different SVG-related tasks.

To better accommodate the unique elements of SVG code, we augment the original Qwen3-VL tokenizer with a dedicated set of SVG symbols. These added tokens are designed to absorb frequent multi-character patterns that would otherwise be fragmented into long subword sequences. As listed in Table~\ref{tab:svg-vocab}, the vocabulary includes 49 tag-level tokens that cover common structural and graphical elements, such as \verb|<svg|, \verb|<path|, and \verb|<circle|. We further introduce 35 attribute-level tokens for geometric fields and style declarations, including tokens like \verb|stroke="|, \verb|class="|, \verb|d="|, and \verb|fill="|.

In addition to symbolic tokens, we explicitly allocate numeric tokens for compact coordinate and parameter prediction. Specifically, the vocabulary contains 247 integer tokens ranging from $-128$ to $128$, together with 100 two-decimal fractional tokens and 10 one-decimal fractional tokens. This numeric design allows the model to express geometric quantities more directly, while shortening output sequences and reducing decoding overhead.

The embeddings of the newly added tokens are initialized using a subword-based initialization strategy. For each added token, we first decompose it using the original tokenizer and then set its initial embedding as the average of the corresponding subword embeddings. This strategy provides a smooth initialization for all SVG-specific tokens, preserving compatibility with the pretrained embedding space. After expansion, all parameters are optimized jointly in an end-to-end fashion. In practice, this initialization scheme improves training stability and helps the model more reliably generate structurally valid SVG code and numerically accurate parameters.

\begin{table}[htbp]
  \centering
  \footnotesize
  \setlength{\tabcolsep}{6pt}
  \renewcommand{\arraystretch}{0.95}
  \captionsetup{justification=centering}
  \caption{SVG-specific tokens used in \method.}
  \label{tab:svg-vocab}

  \begingroup
  \setlength{\arrayrulewidth}{0.5pt}

  \captionsetup{justification=centering}
  \subcaption*{(a) Tag tokens}
  \begin{tabularx}{\textwidth}{>{\centering\arraybackslash}p{0.22\textwidth}|>{\ttfamily\arraybackslash}X}
    \hline
    \textbf{Category} & \textbf{Tokens} \\
    \hline
    Root
      & <svg, </svg>, <defs, </defs>, <use, </use> ,/>\\
    \hline
    Grouping
      & <g, </g> \\
    \hline
    \vspace{0pt}\centering
    Shapes
      & <path, </path>, <rect, </rect>, <circle, </circle>, <ellipse, </ellipse>, <line, </line>, <polyline, </polyline>, <polygon, </polygon> \\
    \hline
    \vspace{0pt}\centering
    Text
      & <text, </text>, <tspan, </tspan>, <textPath, </textPath> \\
    \hline
    \vspace{0pt}\centering
    Gradients
      & <linearGradient, </linearGradient>, <radialGradient, </radialGradient>, <stop, </stop> \\
    \hline
    Clipping
      & <clipPath, </clipPath>, <mask, </mask> \\
    \hline
    \vspace{0pt}\centering
    Filters
      & <filter, </filter>, <feGaussianBlur, </feGaussianBlur>, <feColorMatrix, </feColorMatrix>, <feComposite, </feComposite>, <feBlend, </feBlend> \\
    \hline
  \end{tabularx}

  \vspace{0.6em}

  \subcaption*{(b) Attribute tokens}
  \begin{tabularx}{\textwidth}{>{\centering\arraybackslash}p{0.22\textwidth}|>{\ttfamily\arraybackslash}X}
    \hline
    \textbf{Category} & \textbf{Tokens} \\
    \hline
    \vspace{0pt}\centering
    Geometry
      & width=", height=", viewBox=", x=", y=", x1=", y1=", x2=", y2=", cx=", cy=", r=", rx=", ry=", d=", points=" \\
    \hline
    \vspace{0pt}\centering
    Styling
      & fill=", stroke=", stroke-width=", stroke-linecap=", stroke-linejoin=", stroke-miterlimit=", fill-rule=", opacity=" \\
    \hline
    Transform
      & transform=" \\
    \hline
    Text
      & font-size=", font-family=", text-anchor=" \\
    \hline
    \vspace{0pt}\centering
    Gradients
      & gradientUnits=", gradientTransform=", offset=", stop-color=" \\
    \hline
    Identifiers
      & id=", class=", clip-path=" \\
    \hline
  \end{tabularx}

  \endgroup
\end{table}

\clearpage
\subsection{Examples of SVG-Sophia}

\begin{figure}[htbp]
    \centering
    \includegraphics[width=\linewidth]{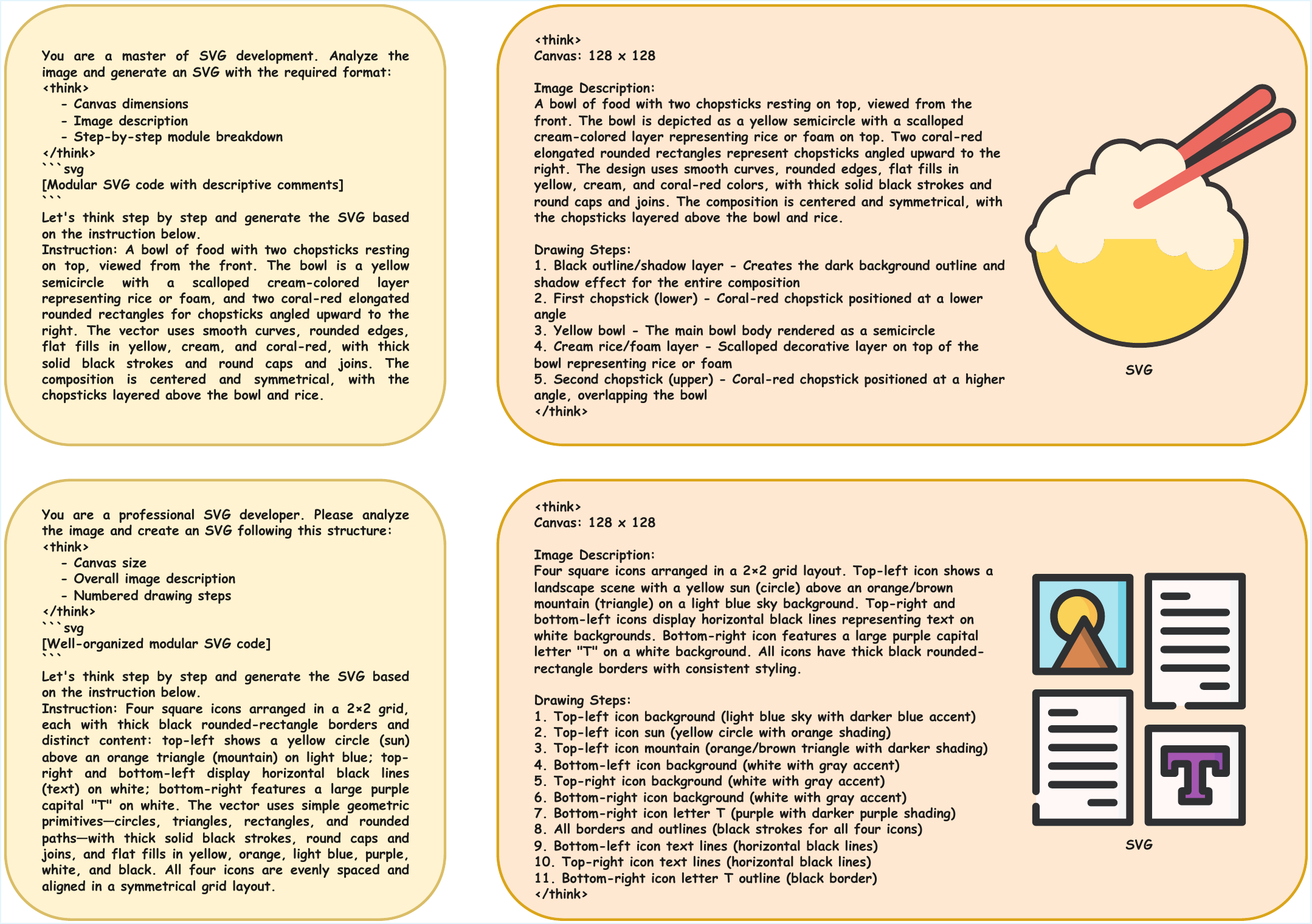}
    \captionsetup{justification=centering}
    \caption{Examples of Text-to-SVG in SVG-Sophia.}
    \label{fig:t2s_appendix_demo}
\end{figure}

\begin{figure}[htbp]
    \centering
    \includegraphics[width=\linewidth]{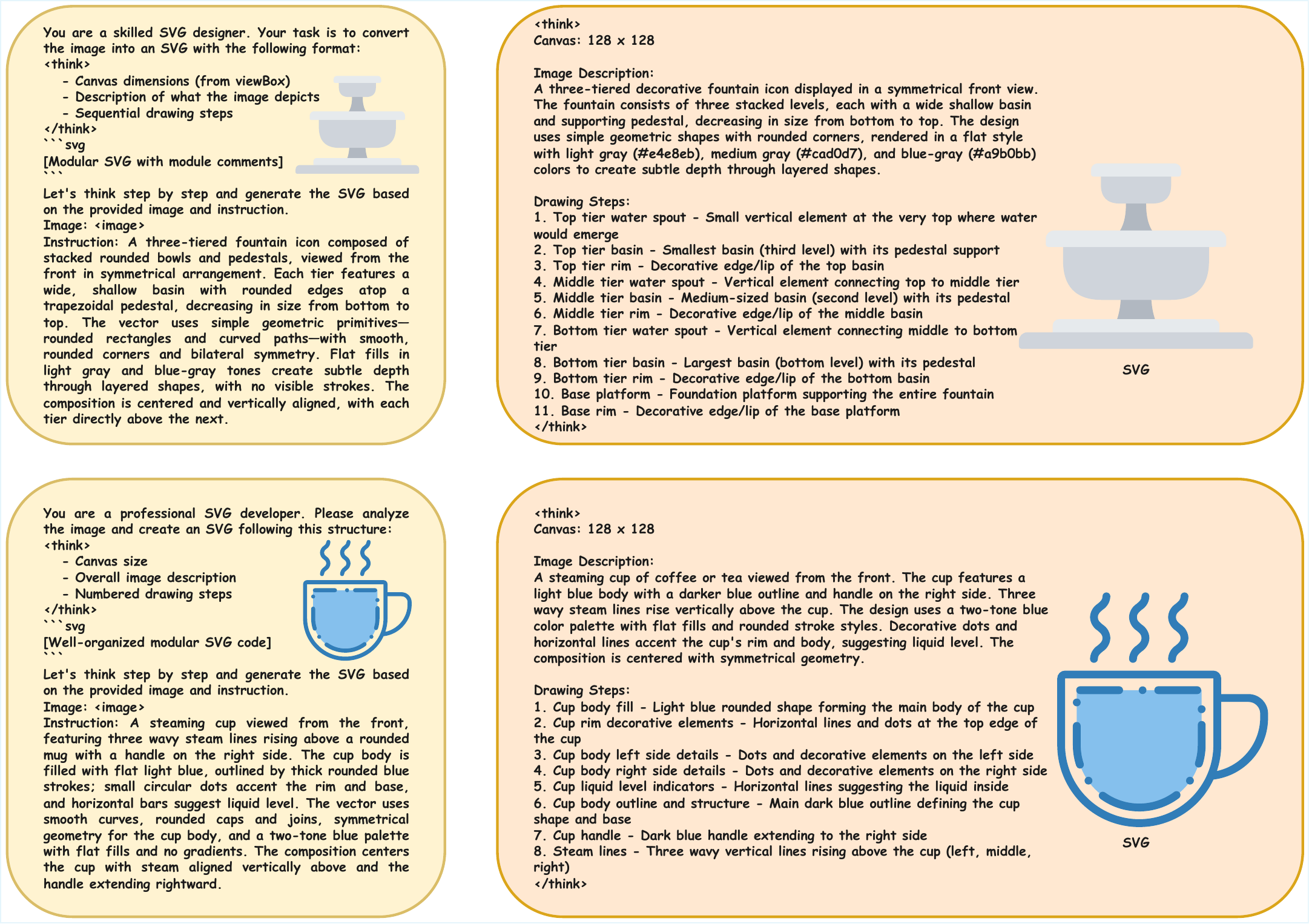}
    \captionsetup{justification=centering}
    \caption{Examples of Image-to-SVG in SVG-Sophia.}
    \label{fig:i2s_appendix_demo}
\end{figure}

\begin{figure}[htbp]
    \centering
    \includegraphics[width=\linewidth]{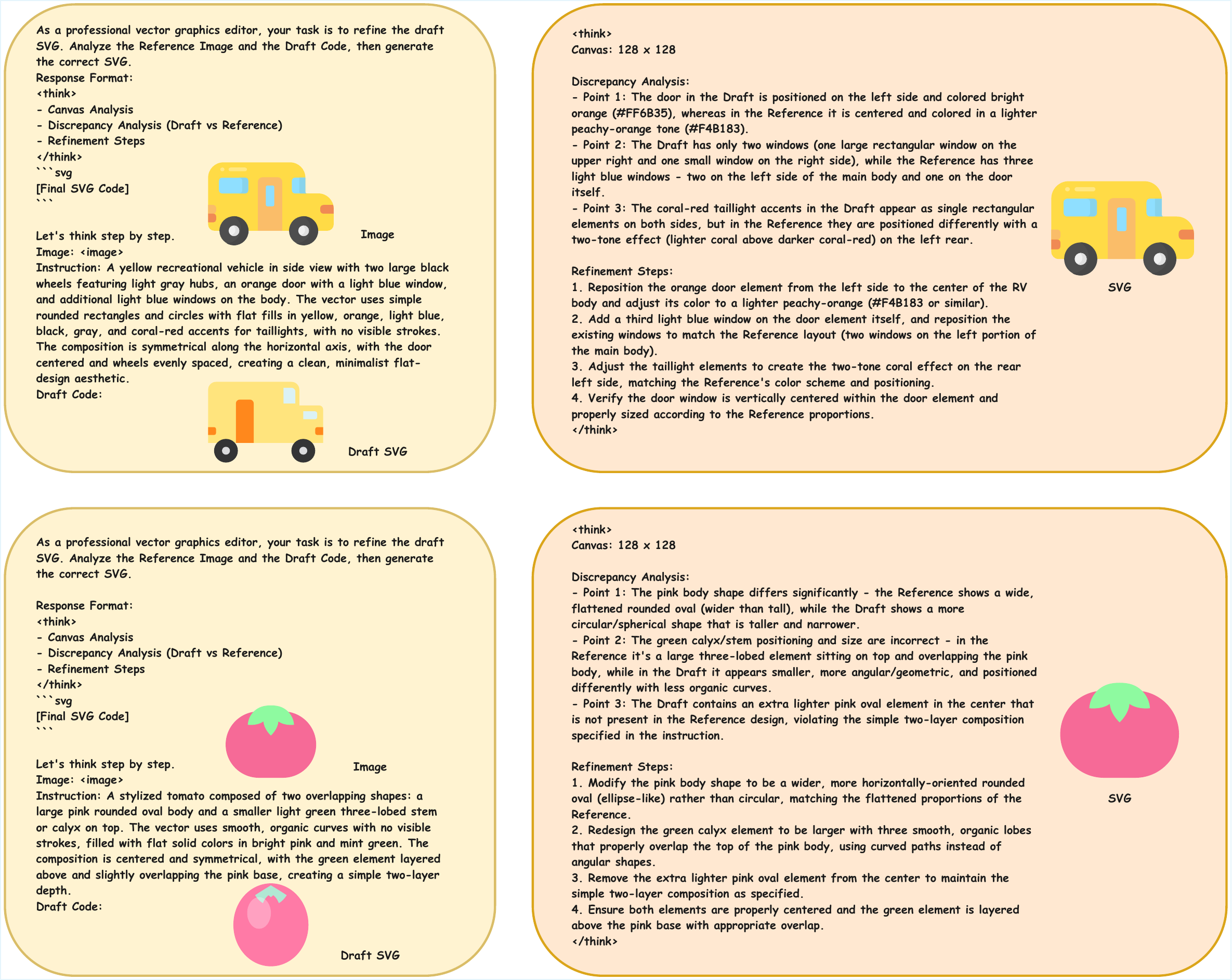}
    \captionsetup{justification=centering}
    \caption{Examples of SVG code refinement in SVG-Sophia.}
    \label{fig:refine_appendix_demo}
\end{figure}

To provide a more intuitive understanding of SVG-Sophia, this section presents representative examples of Text-to-SVG, Image-to-SVG, and SVG code refinement tasks. Specifically, for the Text-to-SVG task, we provide detailed instructions to guide the model in generating high-quality SVGs. For the Image-to-SVG task, we supplement these instructions with raster images to facilitate accurate vectorization. For the SVG code refinement task, we supply the rasterized target image, instructions, and the currently defective SVG code to guide the model in code repair. As illustrated in Figures~\ref{fig:t2s_appendix_demo}, \ref{fig:i2s_appendix_demo}, and \ref{fig:refine_appendix_demo}, each sample features structured reasoning alongside its corresponding SVG code, demonstrating the diverse visual concepts and code patterns covered by SVG-Sophia. These examples highlight the high-quality annotations within SVG-Sophia and showcase how the dataset facilitates both structured reasoning and fine-grained SVG generation and code refinement.

\end{document}